\documentclass{sigchi}

\toappear{Preprint to appear in \emph{HAI 2017, Bielefeld, Germany.}} 

\usepackage{balance}       
\usepackage{graphicx}      
\usepackage[T1]{fontenc}   
\usepackage{txfonts}
\usepackage{mathptmx}
\usepackage[pdflang={en-US},pdftex]{hyperref}
\usepackage{color}
\usepackage{booktabs}
\usepackage{textcomp}
\usepackage{multicol}

\usepackage{microtype}        
\usepackage{ccicons}          

\usepackage{todonotes}

\def\plaintitle{Regulating Highly Automated Robot Ecologies:\\Insights from Three User Studies}

\def\emptyauthor{}
\def\plainkeywords{}

\makeatletter
\def\url@leostyle{%
  \@ifundefined{selectfont}{
    \def\UrlFont{\sf}
  }{
    \def\UrlFont{\small\bf\ttfamily}
  }}
\makeatother
\def\pprw{8.5in}
\def\pprh{11in}

\definecolor{linkColor}{RGB}{6,125,233}
\hypersetup{%
  pdftitle={\plaintitle},
  pdfauthor={\emptyauthor},
  pdfkeywords={\plainkeywords},
  pdfdisplaydoctitle=true, 
  bookmarksnumbered,
  pdfstartview={FitH},
  colorlinks,
  citecolor=black,
  filecolor=black,
  linkcolor=black,
  urlcolor=linkColor,
  breaklinks=true,
  hypertexnames=false
}

\begin{document}

\title{\plaintitle}

\numberofauthors{4}
\author{%
  \alignauthor{Wen Shen\\
    \affaddr{University of California, Irvine}\\
    \affaddr{Irvine, CA, USA}\\
    \email{wen.shen@uci.edu}}
%
%
%
  \alignauthor{Alanoud Al Khemeiri\\
    \affaddr{Masdar Institute}\\
    \affaddr{Abu Dhabi, UAE}
    \email{aalkhemiri@gmail.com}}\\
  \alignauthor{Abdulla Almehrzi\\
    \affaddr{Masdar Institute}\\
    \affaddr{Abu Dhabi, UAE}
    \email{aa.almehrezi@gmail.com}}
  \alignauthor{Wael Al Enezi\\
    \affaddr{Masdar Institute}\\
    \affaddr{Abu Dhabi, UAE}\\
    \email{wralenezi@gmail.com}}
  \alignauthor{Iyad Rahwan\\
    \affaddr{MIT Media Lab}\\
    \affaddr{Cambridge, MA, USA}\\
    \email{irahwan@mit.edu}}
  \alignauthor{Jacob W. Crandall\\
    \affaddr{Brigham Young University}\\
    \affaddr{Provo, UT, USA}\\
    \email{crandall@cs.byu.edu}}
}


\maketitle

\begin{abstract}
Highly automated robot ecologies (HARE), or societies of independent autonomous robots or agents, are rapidly becoming an important part of much of the world's critical infrastructure.  As with human societies, regulation, wherein a governing body designs rules and processes for the society, plays an important role in ensuring that HARE meet societal objectives.  However, to date, a careful study of interactions between a regulator and HARE is lacking.  In this paper, we report on three user studies which give insights into how to design systems that allow people, acting as the regulatory authority, to effectively interact with HARE.  As in the study of political systems in which governments regulate human societies, our studies analyze how interactions between HARE and regulators are impacted by regulatory power and individual (robot or agent) autonomy.  Our results show that regulator power, decision support, and adaptive autonomy can each diminish the social welfare of HARE, and hint at how these seemingly desirable mechanisms can be designed so that they become part of successful HARE.
\end{abstract}




\section{Introduction}

Centuries of political discourse have led to diverse philosophies for how to best govern human societies.  Political opinions advocate everything from highly controlled societies (authoritarianism) to loosely controlled societies (e.g., libertarianism), and just about everything in between.  Political philosophies differ with respect to the extent of power and resources given to governments, as well as the rights and autonomy of individuals in society~(e.g., \cite{Hsu2008,Locke1689,Rand1957}). 

Similar discourse is necessary in the context of highly automated robot ecologies (HARE), which are collections of independent and autonomous robots, agents, or software systems that share constrained resources.  For example, it is not hard to imagine future transportation systems composed almost entirely of independent driverless cars.  Other aspects of modern cities, including smart grids and smart buildings, consist of networks of autonomous robotic devices that compete for and share (potentially constrained) water and electricity resources.  Similarly, investor behavior in financial markets is increasingly driven by sophisticated control algorithms.  As in the governing of human societies, regulators (or {\em regulatory authorities}) consisting of one or more people are given resources and power to influence the behavior of these HARE.  The goal of these human-agent interactions is to ensure that shared resources are effectively and appropriately utilized.  

Despite similarities between regulating human and robot societies, there are also glaring differences, not the least of which is that individual and collective robot behavior is often quite distinct from human behavior.  Robots and other AIs can sometimes respond to stimuli instantaneously and in mass in ways that people cannot~\cite{Kirlienko2016}.  Likewise, robots are not likely to respond identically to regulations (e.g., information, incentives, and force) as people, nor might they be afforded the same rights.  Thus, given the rapid rise of HARE in modern critical infrastructure, it is important that we study interactions between HARE and regulatory authorities in order to design systems that meet societal objectives.  

In this paper, we study, via three user studies, how to design systems that allow people, acting as the regulatory authority, to effectively govern HARE.  As in the study of political systems, our studies analyze, under an initial set of assumptions, how simple HARE are impacted by regulatory power and individual (robot) autonomy.  Results show that regulator power, decision support, and adaptive robot autonomy can each diminish the social welfare of the HARE in some conditions, and suggest how these seemingly desirable mechanisms can be used so that they become part of more successful HARE.

While these user studies and the associated analysis do not (and, indeed, cannot) provide universal or general statements about all HARE, the intended contribution of this paper is to raise awareness of the potential pitfalls and opportunities that should be considered in the design of real-world HARE.


\section{Interacting with HARE}

Before describing the user studies, we discuss HARE and the regulator authority's role in interacting with HARE.

\subsection{HARE}

A highly automated robot ecology (HARE) is a collection of independent and autonomous robots that either share resources or participate in the same activity.  Both the terms {\em independent} and {\em autonomous} deserve explanations.  The robots are {\em independent} from each other in that they are owned by different stakeholders.  No one person or organization owns all robots in the collective, as individual stakeholders decide the goals and algorithms used by their robots.  This independence in ownership and design (and, hence, goals and algorithms) does not imply that the robots do not impact each other.  The robots may communicate with each other.  Furthermore, each robot's environment is impacted by the other robots' actions.

In a HARE, the robots are {\em autonomous} in that, from the regulator's perspective, they make their own decisions.  A regulatory authority cannot interrupt or override the robots' decision-making algorithms without the permission of their stakeholders.  However, the regulator may change the robots' environment (by supplying information, providing incentives, changing physical infrastructure, etc.) to influence them.

Each robot's behavior is determined by its control algorithm.  Financial incentives and other objectives often drive stakeholders to equip their robots with sophisticated and adaptive control algorithms~\cite{amin2000,wang2008,mitchell2004} designed to maximize the individual stakeholders' benefits rather than societal objectives.  As such, collective behavior often fails to meet societal goals.  The extent to which HARE fall short of desirable societal outcomes is known as {\em the price of anarchy}~\cite{PriceofAnarchy,may2010, heldane2011}.

\subsection{Regulation}

Since the price of anarchy can be quite high~\cite{Youn2008}, regulatory authorities are established to set rules and incentives that promote system-wide stability and efficiency.  For example, a transportation authority assigned to regulate driverless cars can use road structure, information, and penalties and incentives (e.g., tolls or ticketing) to promote efficient and safe traffic flow.  Regulators of new-age power systems can use contracts~\cite{contracts2009}, information-based interventions \cite{Schultz2007}, and real-time pricing \cite{Borenstein2002} to influence robotic buildings to reduce peak consumption and match electricity demand to supply.  In each case, the regulatory authority can potentially reduce the price of anarchy by altering the robots' environment.

Interactions between regulatory authorities and HARE bring to mind mechanism design~\cite{Hurwicz2006}, supervisory control (SC) of multiple robots~\cite{chen2011,zheng-glas-tcms2014}, and systems with shared autonomy~\cite{CrandallIROS2002,Johnson2012,Nikolaidis2017}.  These research areas have both key similarities and differences with regulating HARE.  We discuss each in turn.

\subsubsection{Relation to Mechanism Design}
The regulatory authority engineers the environment to encourage cooperation among robots in the HARE.  Cooperation is most easily achieved by either influencing the robots to converge to a more efficient equilibrium or to alter the scenario so that it has a unique, more efficient equilibrium.  This later problem, called
{\em mechanism design}~\cite{Hurwicz2006} (i.e., reverse game theory), has been applied to many domains including power grids~\cite{vytelingum2010trading}, financial markets~\cite{mackie2006automated} and transportation systems~\cite{zhang2016control,shen2016online}.

The goal of mechanism design is to implement strategy-proof mechanisms (e.g., payment schemes) that incentivize agents to truthfully reveal their private information~\cite{Hurwicz2006,nisan2007algorithmic}.  Unfortunately, such mechanisms do not always exist, especially in online and dynamic settings~\cite{nisan2007algorithmic,pavan2009dynamic,parkes2004mdp}.  Furthermore, computational complexity and privacy concerns often prohibit incentive-compatible mechanisms from being implemented even when they do exist in theory~\cite{mcsherry2007mechanism,nisan1999algorithmic,feigenbaum2002distributed}.  In HARE, necessary prior knowledge required to implement such mechanisms (e.g., the robots' state and action spaces)~\cite{Hurwicz2006,nisan2007algorithmic}, is often not immediately available, and it is often not possible to obtain this information through auctions or similar revealed-preference mechanisms in a timely fashion~\cite{myerson1981optimal,trigo2011collective,chambers2016revealed}. Therefore, the regulator must experiment in real time to identify interventions that produce desirable societal outcomes.

\subsubsection{Relation to Supervisory Control}
Human-HARE interactions also call to mind traditional supervisory-control (SC) systems in which an operator directs multiple (semi-autonomous) robots (e.g., \cite{chen2011,zheng-glas-tcms2014}).  While we anticipate that these operators face similar challenges as regulators of HARE (e.g., situation awareness~\cite{Endsley1988} and operator workload), there are critical differences.  For example, robots in HARE are autonomous (level 10) from the regulator's perspective, while robots in SC systems typically operate at a lower level of automation~\cite{SheridanVerplank1978}.  Thus, in SC, operators can directly override or alter the robots' decision-making, algorithms, or goals.  Regulators of HARE cannot.

\subsubsection{Relation to Shared Autonomy}
In HARE, system dynamics are governed by the behavior of both the regulator and the robots, consistent with the idea of {\em shared autonomy} (also called shared control~\cite{CrandallIROS2002}).  One particularly relevant application of shared control~\cite{Crandall2017} that has parallels to regulating HARE is human-swarm interaction (HSI)~\cite{kolling2013human,BrownJHRI_2016}.  In HSI, an operator commands or influences a set of robots that have been programmed to mimic biological swarms.  To do this, each robot in the swarm is equipped with simple known (to the operator) control algorithms.  Despite similarities, HARE differ from traditionally defined robot swarms in that robots in HARE are programmed by separate stakeholders.  Thus, the algorithms are not likely to be known to the regulator, may be highly sophisticated, and are not guaranteed to be the same among all robots.  As a result, organized, cooperative group behavior can be more difficult to achieve in HARE than in traditional robot swarms, and different interactions are likely necessary.

\subsection{Parameters}

Given differences between regulating HARE and other better-studied systems, we seek to understand how and when HARE can be effectively regulated.  While HARE can be parameterized in many ways (including the frequency of decision-making~\cite{vespignani2009predicting,johnson2013abrupt} and the switching processes of system states~\cite{preis2011switching}), we study two important attributes in this paper: {\em regulatory power} and {\em robot control algorithms}.  


The jurisdiction and resources given to the regulatory authority to carry out its intended functions define {\em regulatory power}.  Regulatory power determines the interventions the regulator can use.  For example, local laws determine whether a transportation authority is allowed to charge tolls, how or in what manner it can change tolls, and how it can enforce payment.  Furthermore, monetary resources impact which toll systems it is able to implement and maintain.  Similarly, a utility company seeking to modulate the behavior of robotic buildings is limited by laws and resources that govern, among other things, the information the utility can collect and the pricing incentives it can successfully implement.


The control algorithms employed by individual robots also play an important role in HARE.  Algorithms differ along many dimensions, including the data sources utilized by the algorithm, the algorithm's depth of reasoning, and the algorithm's adaptivity.  In this paper, we consider how the ability of people to effectively regulate HARE is impacted by the ability of the robots to learn from past experiences.  We refer to algorithms that do not adapt as {\em simple automation}, and to those that do adapt based on past experience as {\em adaptive automation}.

\subsection{Assumptions}

Regulating HARE is a rather vast topic that we cannot fully address in a single work.  For simplicity, we assume that regulators use monetary incentives to influence robot behavior and the regulatory authority consists of a single person.  We also work with simulated environments (environments with simplified dynamics but which maintain many of the important characteristics of real-world HARE) to simplify data gathering.  Though not without limitations, these simplifications offer a reasonable starting point to study various aspects of regulating HARE.  Future work can and should relax these simplifications.




Given these assumptions, we begin to evaluate how regulatory power and algorithm adaptivity impact people's ability to regulate HARE via a series of user studies.

\section{Regulation and Adaptivity}

To begin to understand how regulatory power and robot adaptivity jointly impact people's abilities to regulate HARE, we conducted two user studies in which participants regulated simulated HARE.  In the first study, participants used tolls to manage a simple transportation system composed of autonomous driverless cars.  In the second study, participants regulated robotic buildings that shared a limited water supply.   Both studies were 2x3 between-subjects designs in which we varied robot adaptivity and regulatory power.

\subsection{User Study 1 -- Driverless Cars}
We study a HARE composed of simulated driverless cars.  

\begin{figure}[t]
\centering
\includegraphics[height=1.5in]{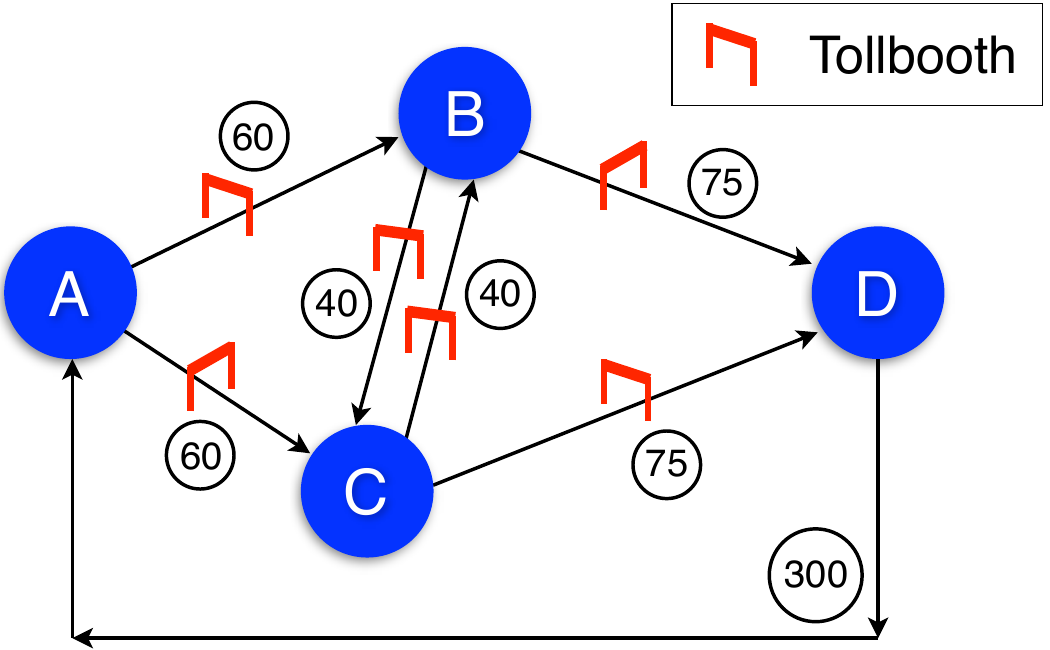}
\caption{A transportation network represented by a directed graph.  Circled numbers specify road capacity.}
\label{fig:trans_graph}
\end{figure}

\subsubsection{Scenario Overview}
Simulated autonomous cars used routing algorithms to navigate through a simple transportation network (Figure~\ref{fig:trans_graph}).  Cars traveled at velocities determined by the number of vehicles on a road.  When traffic was below a road's capacity, cars moved at maximum speed.  But when traffic approached and exceeded the road's capacity, traffic flow slowed to a crawl (see Appendix~A for details).

\begin{figure}[t]
\centering
\includegraphics[width=0.38\textwidth]{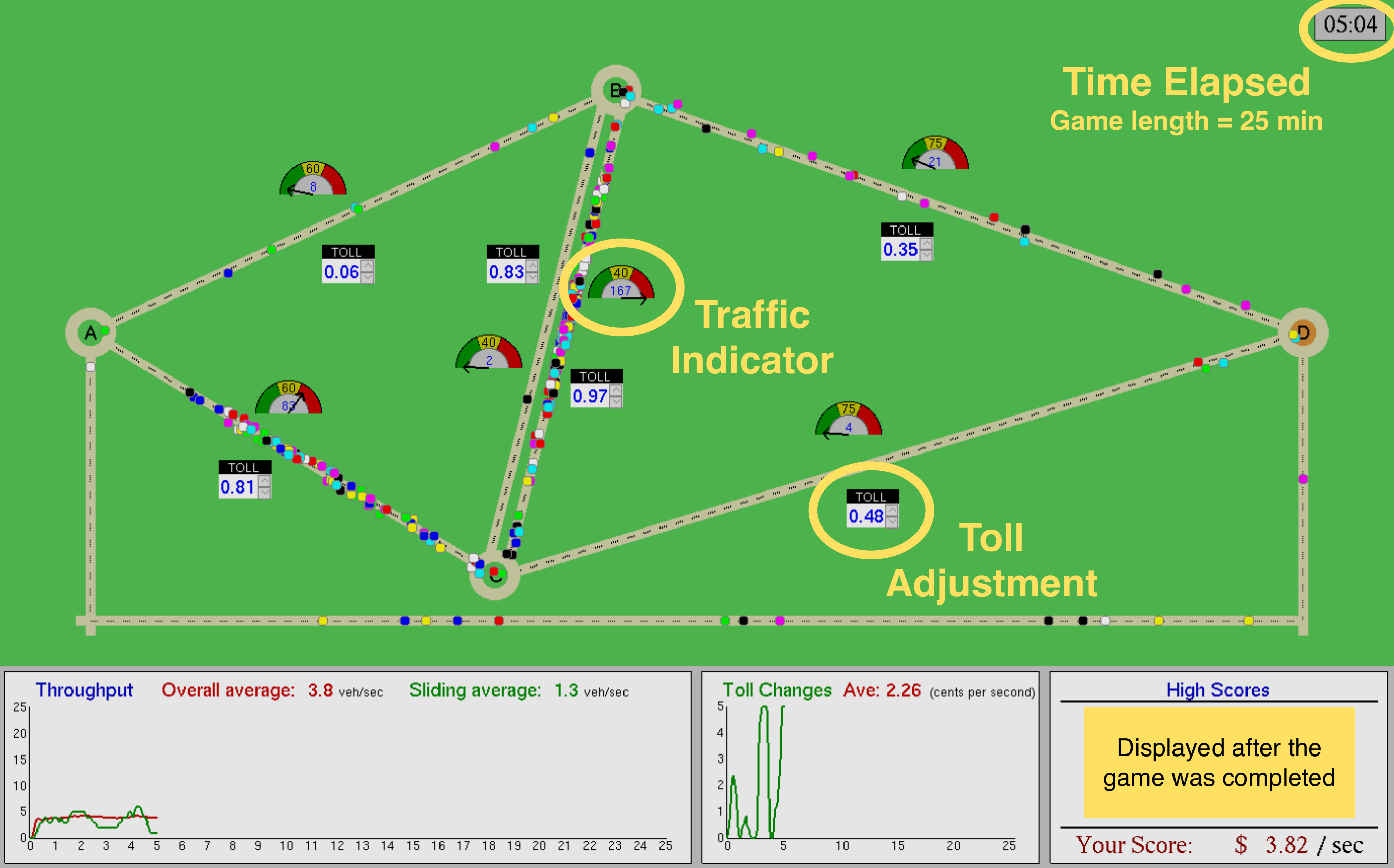}
\caption{The GUI used in user study 1.  Annotations (yellow) are overlaid for explanatory purposes.}
\label{fig:jiaotong}
\end{figure}

The regulatory authority was tasked with regulating the driverless cars so as to maximize traffic flow through the network, which was measured as the throughput through node~D.  To influence the cars, the regulatory authority set tolls on each road using a GUI (Figure~\ref{fig:jiaotong}) showing a bird's-eye view of the transportation network, including the current location of each of the 300 cars.  The GUI also displayed the number of cars currently on each road, as well as each road's capacity.  Toll changes were announced instantaneously to all cars.  

Initially, tolls on all roads were set to \$0.50.  Participants could increase or decrease each toll (between \$0.00 and \$0.99) by clicking on the corresponding buttons.  Thus, if road $BC$ was overcrowded, a regulator might consider trying to reduce the traffic congestion on this road by increasing the toll on $BC$, decreasing the toll on $BD$, increasing the toll on $AB$, decreasing the toll on $AC$, or using some combination of these methods. By properly balancing the various tolls, the regulator could eliminate congestion, which in turn produced high throughput through node~D.  Participants could click the buttons in rapid succession to quickly make large toll changes.

Each robot continually moved through the transportation network, repeatedly selecting a destination node and a route to that node from its current location so as to maximize its own utility.  A car received positive utility each time it arrived at its destination, but incurred costs for tolls incurred and a (per unit time) operational cost.  Thus, routes expected to take longer to traverse or that had higher tolls tended to yield lower utility and were more likely to be avoided by the cars.  

Formally, each car estimated its current utility for going to destination $g$ from its current location $i$ as follows:
\begin{eqnarray}
u(i,g) = v(g) - c_t(i,g) - c_\$(i,g), 
\label{eq:utility}
\end{eqnarray}
where $v(g)$ was the utility for arriving at destination $g$, $c_t(i,g)$ was the estimated travel cost for going from the car's current location to destination $g$, and $c_\$(i,g)$ was the projected toll charge for going to destination $g$ (see Appendix B for details).

Since neither $v(g)$ nor $c_t(i,g)$ (and how they might compare to $c_\$(i,g)$) were known to the regulator (for any car), the regulator could only determine how tolls might impact the cars' behavior through experimentation and observation.

%
%

\subsubsection{Experimental Setup}
\label{sec:carexpsetup}
We conducted a user study in which people regulated 300 simulated cars.  In this study, we varied both {\em robot adaptivity} and {\em regulatory power} to determine how these two variables jointly impact people's ability to effectively regulate HARE.  As summarized in Table~\ref{tab:algsoph}, robot adaptivity contained two factor levels indicating the type of navigation system used by all the cars: simple automation and adaptive automation.  In both cases, each car used Dijkstra's Algorithm and Eq.~(\ref{eq:utility}) to determine which path to follow.  However, the cars used different mechanisms to estimate travel costs ($c_t(i,g)$).  Cars that used simple automation estimated travel costs assuming a congestion-free network.  On the other hand, cars that used adaptive automation estimated travel costs on each road using reinforcement learning (Appendix B).  Thus, cars that used simple automation did not learn from their past experiences (and, hence, only reacted to toll changes), whereas cars that used adaptive automation learned over time.

\begin{table}[t!]
\caption{Factor levels for robot adaptivity, which were defined based on how travel costs ($c_t(i,g)$) were estimated (Appendix B).}
\label{tab:algsoph}
\begin{center}
\begin{tabular}{l|l} \hline
{\bf Level} & {\bf Cost Estimation} \\ \hline
Simple & {\small The cars did not learn from their past experiences.} \\
automation &  {\small Travel costs were estimated assuming no congestion.} \\ \hline
Adaptive & {\small All cars used reinforcement learning (based on their} \\
automation & {\small own experiences) to determine travel costs.} \\ \hline
\end{tabular}
\end{center}
\end{table}

We considered three levels of regulator power: none, limited and unlimited (Table~\ref{tab:regpow}).  For no regulatory power, no toll changes were permitted (no participants needed).  When given unlimited regulatory power, participants could change tolls as frequently and as much as they desired.  However, under limited regulatory power, participants were given a budget which limited the total amount of toll changes.  Initially, participants received a toll-change fund of \$0.30, which increased by \$0.007 each second.  Thus, the total toll-change budget for a 25-minute game was \$10.80.  The absolute value of each toll change was subtracted from the budget.  Toll changes were not permitted that caused the budget to drop below zero.

\begin{table}[t!]
\caption{Factor levels for regulatory power.}
\label{tab:regpow}
\begin{center}\vspace{-.15in}
\begin{tabular}{l|l} \hline
{\bf Level} & {\bf Description} \\ \hline
None & {\small No toll changes were allowed.} \\ \hline
Limited & {\small Regulators had a budget which limited the amount of toll} \\
& {\small changes they could make.} \\ \hline
Unlimited & {\small Regulators could change tolls as much as they desired.} \\ \hline
\end{tabular}
\end{center}
\end{table}

\subsubsection{Protocol}
\label{sec:protocol1}
Forty-eight students and research staff from Masdar Institute participated in the study.  The following protocol was followed:
\begin{itemize}
  \setlength{\itemsep}{0pt}
  \setlength{\parskip}{0pt}
  \setlength{\parsep}{0pt}
\item The participants were randomly and uniformly assigned across four conditions: Simple-Limited, Adaptive-Limited, Simple-Unlimited, or Adaptive-Unlimited.

\item The participant was trained on how to play the game in the designated condition, but with cars that chose routes randomly.  This training continued until the participant felt comfortable with the objectives of the game, the user interface, and how to set tolls.

\item The participant played a 25-minute game.  Initially, the cars were randomly distributed across the four nodes in the network, which immediately caused congestion to develop on several roads.  The participant needed to bring the system to a congestion-free state as quickly as possible.  Cars were biased so that more cars preferred node C as a destination.  To incentivize high performance, a high-score list was displayed once the game completed.

\item The participant completed a post-experiment questionnaire, which asked which node more cars preferred and whether or not the cars employed learning algorithms.

\end{itemize}\vspace{-.05in}
Twelve trials for both the Simple-None and Adaptive-None conditions were also carried out (no participants required).

\subsubsection{Results}

\begin{figure}[t]
\centering
\vspace{-.12in}
\includegraphics[width=2.6in]{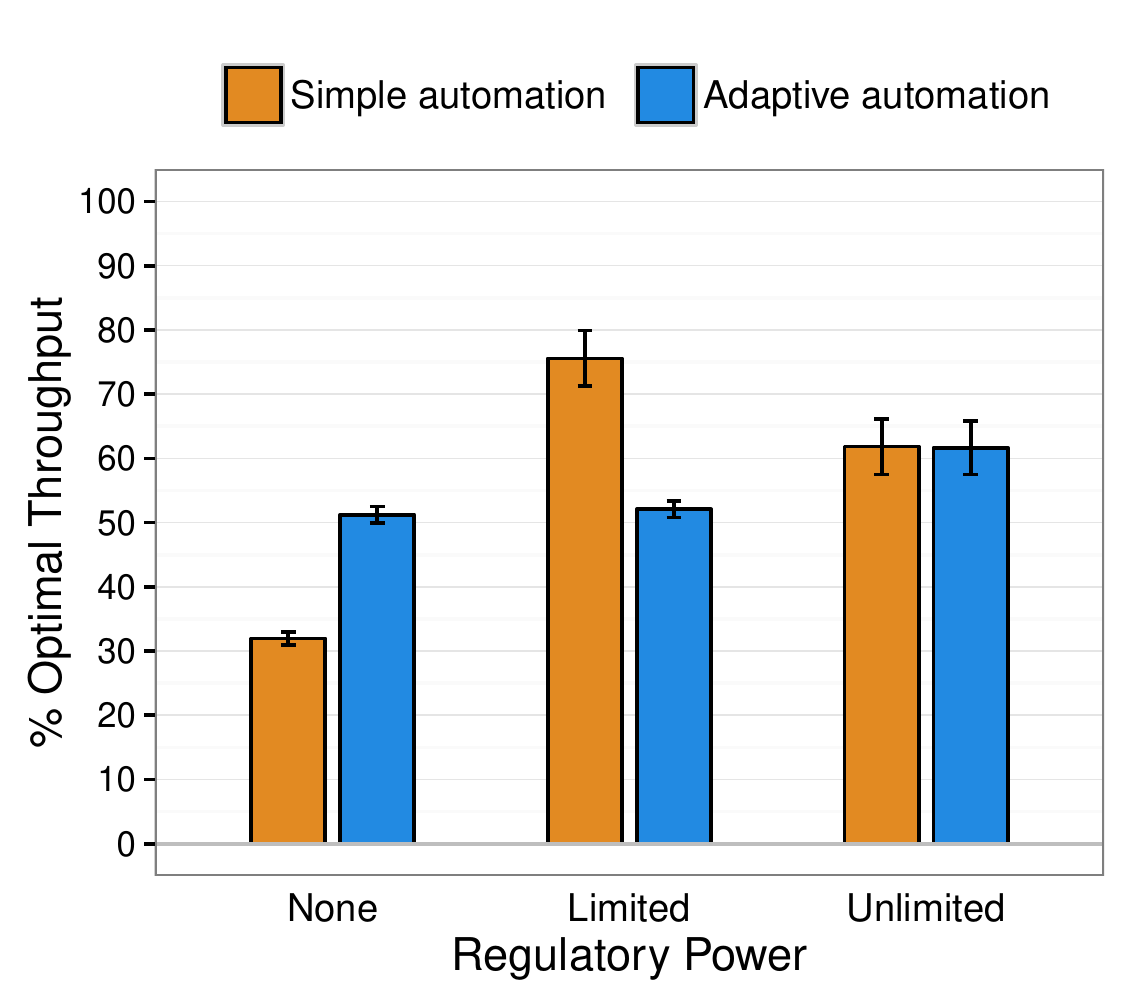} \vspace{-.05in}
\caption{Average throughput observed in user study 1.  Error bars show the standard error of the mean.}
\label{fig:jiaotong_throughput}
\end{figure}

Figure~\ref{fig:jiaotong_throughput} shows the average performance of the HARE, measured as a percentage of optimal throughput over the duration of the game, achieved in each condition.   Absent regulations, societies of driverless cars equipped with adaptive automation performed much better than societies of cars using simple automation.  However, limited regulatory power reversed this trend.  Limited regulatory power led to vastly better outcomes for societies composed of simple robots, but had no impact on societies comprised of adaptive robots.  While additional (unlimited) regulatory power improved the efficiency of adaptive societies by a small amount, it decreased throughput for societies comprised of simple robots.

An analysis of variance, where throughput was the dependent variable and robot adaptivity and regulatory power were independent variables, confirmed many of these trends.  This analysis showed a main affect for regulatory power ($F(1,66)=30.47$, $p < 0.001$), but not for robot adaptivity ($F(2,66)=0.32$, $p = 0.572$).  However, there was an interaction affect between robot adaptivity and regulatory power ($F(2,66)=23.15$, $p < 0.001$).  Tukey post hoc analysis showed that simple automation with no regulation was worse than all other conditions ($p < 0.001$), while simple automation with limited regulatory power was better than all other conditions ($p \leq 0.03$ for each pairing).  Regulatory power had no significant impact on societies of adaptive robots.

\begin{figure}[t]
\centering
\includegraphics[height=1.7in]{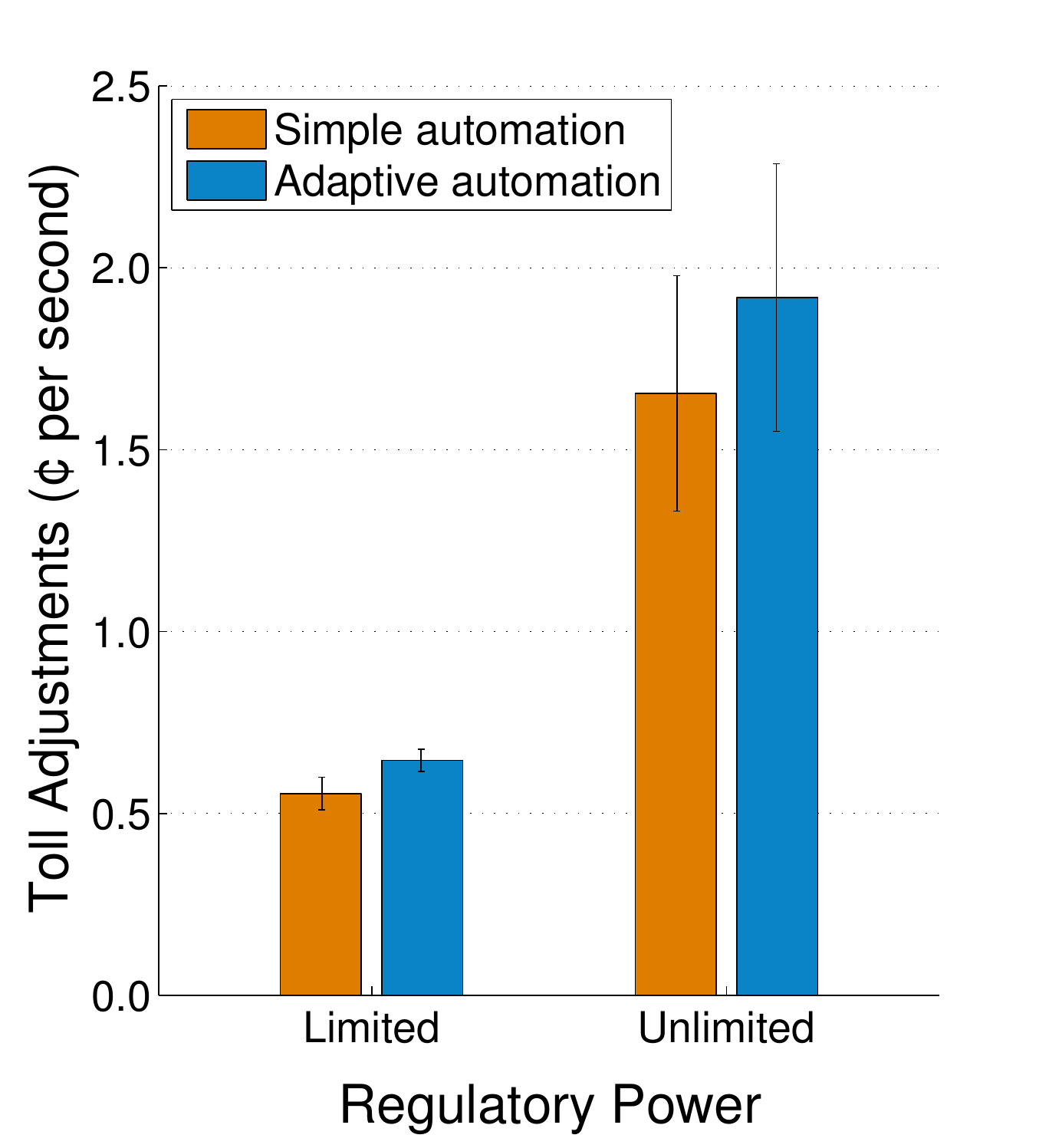}
\includegraphics[height=1.7in]{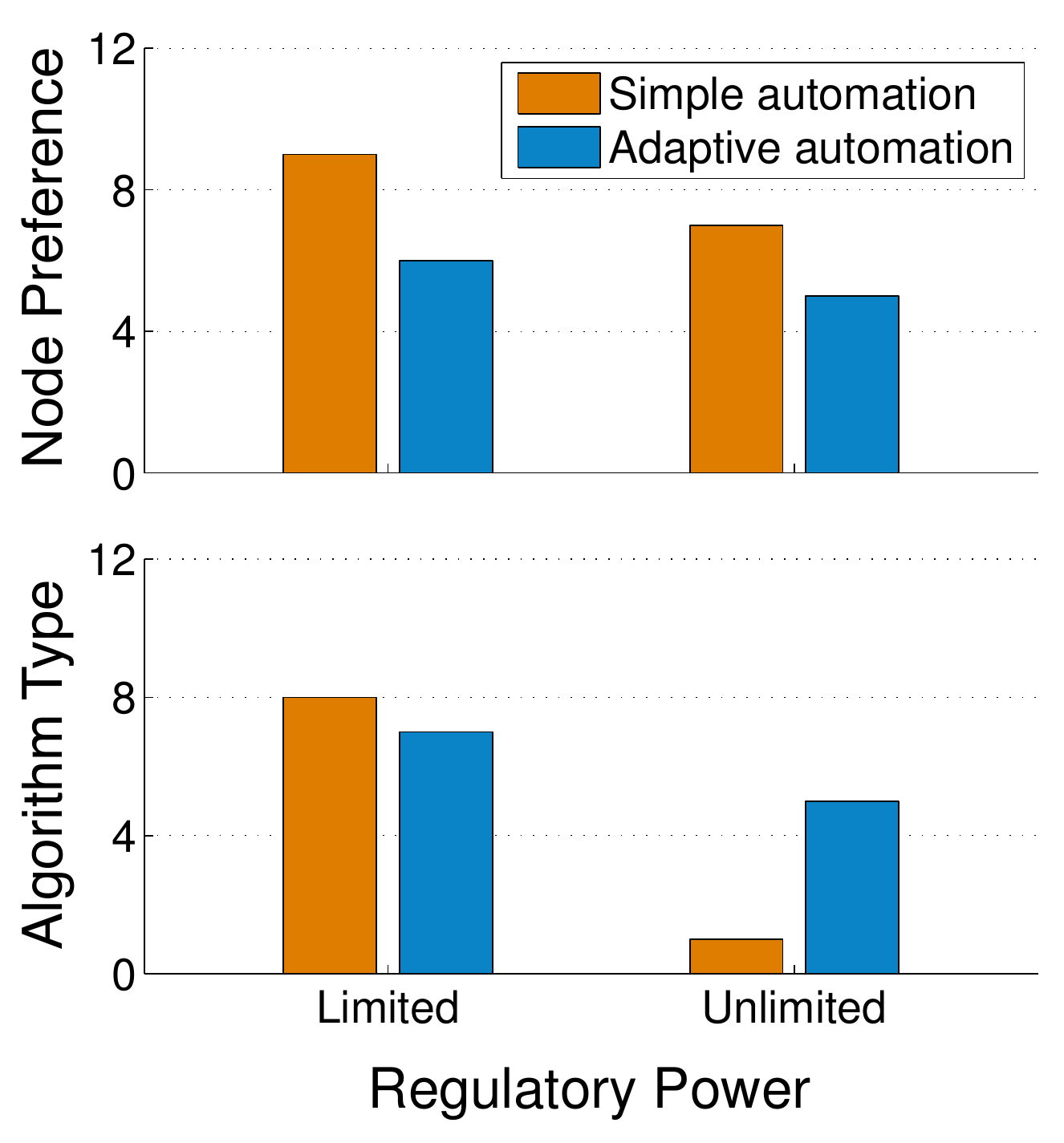}\\
(a) ~~~~~~~~~~~~~~~~~~~~~~~~~~~~~~~~~ (b) ~
\caption{Data from user study 1.  (a) Toll adjustments made per second.  (b) The number of participants that correctly deduced in the post-experiment questionnaire (top) the node most preferred by the cars and (bottom) whether or not the autonomous cars were learning.}
\label{fig:attention_error}
\end{figure}

We attribute the unanticipated drop in performance between the Simple-Limited to the Simple-Unlimited conditions to overuse of regulatory resources, which in turn led to participants having poorer models of the HARE. To see this, consider Figure~\ref{fig:attention_error}a, which shows the amount of toll adjustments made by participants per second in the first user study.  Unsurprisingly, substantially more toll adjustments were made by regulators who had unlimited regulatory power.  While additional toll adjustments may have been justified in the case of adaptive automation, additional interventions were unnecessary when robots used simple automation.  While in the Simple-Limited conditions participants were forced to wait before making more toll changes due to a limited budget, many participants did not do so in the Simple-Unlimited condition.  Rather, they continually made toll adjustments without waiting sufficient time for the robots to adjust~\cite{NeglectBenevolence}.  Thus, they were largely unable to effectively identify which node more robots preferred (Figure~\ref{fig:attention_error}b-top) and whether or not the robots were learning (Figure~\ref{fig:attention_error}b-bottom).  Thus, limited resources appear to have encouraged observation and were, hence, beneficial.

In summary, moderate levels of regulatory power combined with non-adaptive robots had the highest social welfare.  We now consider a second scenario to get a second data point.

\subsection{User Study 2 -- Robotic Buildings}

In this study, participants regulated the activity of tenants in a robotic buildings that shared a limited water supply.

\subsubsection{Scenario Overview}

Eight (simulated) tenants of an apartment building shared a limited water resource.  Each tenant's apartment was equipped with robotic devices that automatically scheduled and executed water-related activities (e.g., laundry, dish-washing, etc.)~on behalf of the tenant.  A tenant programmed its own devices to execute activities automatically using a control algorithm.  Water supplied to the building was collected and purified via a renewable-energy source, a process that limited water availability such that water needs exceeded supply (Appendix~C).

\begin{figure}[t]
\centering
\includegraphics[width=0.38\textwidth]{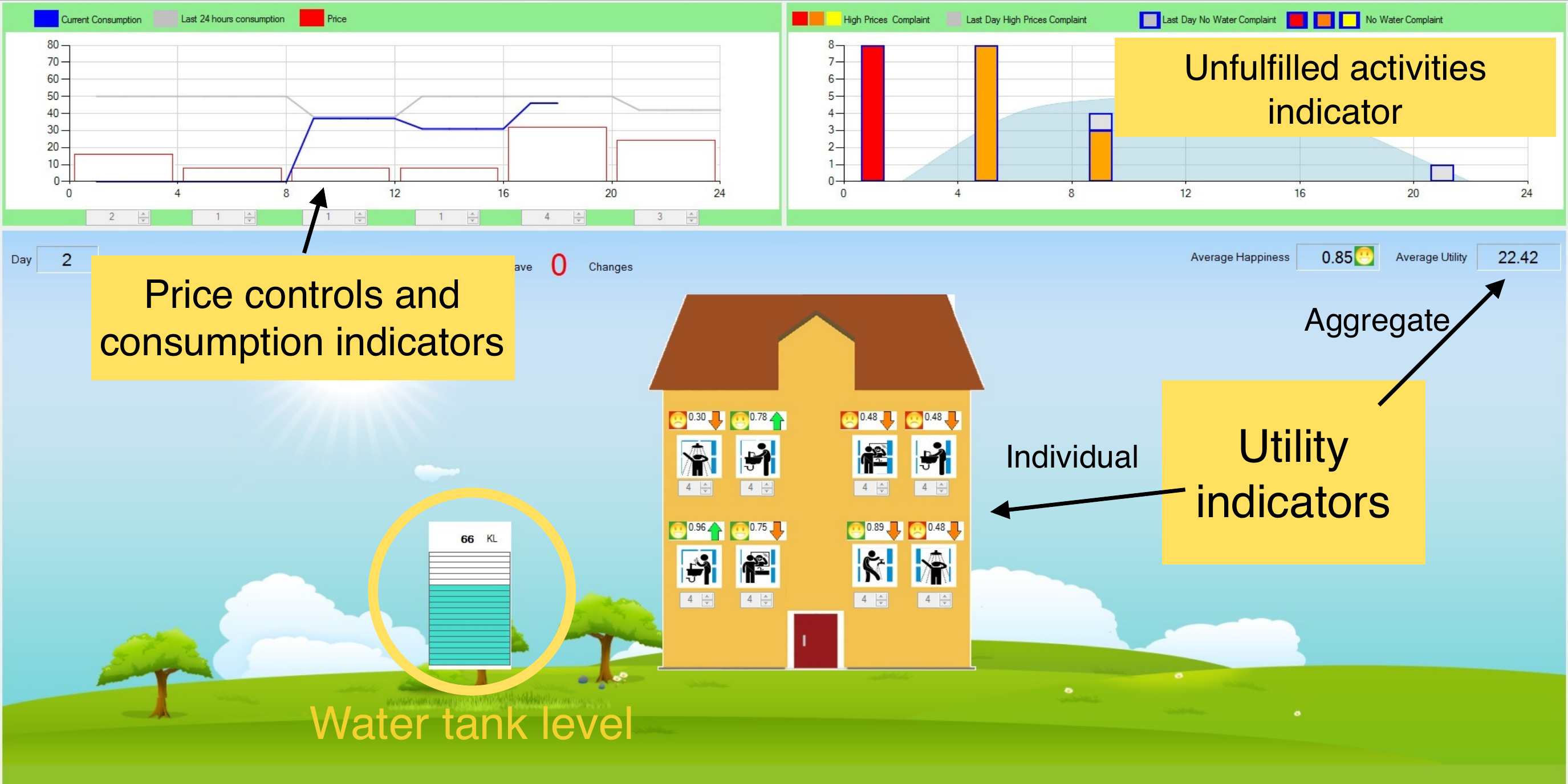}
\caption{The (annotated) GUI used in study 2.}
\label{fig:Atter}
\end{figure}

The regulator's job was to set the per-unit cost of water in each time period (we assumed a day with six time periods) each day such that the aggregate utility across all tenants, days, and periods was maximized.  Participants set prices using the GUI pictured in Figure~\ref{fig:Atter}, which, in addition to allowing participants to change prices, displayed the current water level, the amount of water consumed per period, the number and value of tasks shed by the robotic devices, and the aggregate and individual happiness of the tenants. 

Each tenant employed a control algorithm designed to maximize its total utility.  The water needs of each tenant were defined by a set of activities.  Activity $i$ was defined by the 4-tuple $(t_s(i), t_f(i), s(i), v(i))$, where the time interval $[t_s(i),t_f(i))$ defined the time window during which activity $i$ could be executed, $s(i)$ was the amount of water consumed by activity $i$, and $v(i)$ was how much the tenant valued the completion of activity $i$.  When activity $i$ was carried out, the tenant received utility $u(i) = v(i) - c(i)$, where $c(i)=s(i) p(t)$ was the cost for executing activity $i$ and $p(t)$ was the per-unit cost of water set by the regulator for period $t$.  

Since the tenants' water-related activities (and how the utilities might compare to $c(i)$) were unknown to the regulator, the regulator could only determine what prices to set through experimentation and observation.

\subsubsection{Experimental Setup}
We considered societies in which (1)~devices used simple (non-adaptive) algorithms and (2)~devices used adaptive algorithms to schedule activities. As summarized in Table~\ref{tab:wateralgo}, simple algorithms executed any activity with positive utility when water was available.  They did not adapt their behavior based on their experience.  On the other hand, adaptive algorithms shifted their tenant's activity schedules based on estimates of water availability and price in each time period (Appendix D) to maximize the tenant's expected utility.  We evaluated the same three levels of regulatory power as in Study~1 (Table~\ref{tab:waterfactors}).

\begin{table}[t!]
\caption{Factor levels for robot adaptivity.  See Appendix D for details.}
\label{tab:wateralgo}
\begin{center}
\begin{tabular}{l|l} \hline
{\bf Level} & {\bf Decision Making Process} \\ \hline
Simple & {\small The robotic building carried out activity $i$ if and only if} \\
auto- & {\small $u_{i} > 0$ and there was sufficient water for the activity.}\\ 
mation &  {\small The building did not shift activities based on experience.} \\ \hline
 
Adaptive & {\small The robotic building shifted water-related activities to} \\ 
auto- & {\small maximize its tenant's estimated utilities, which were} \\
mation &  {\small based on estimated hourly prices and water availability.} \\
& {\small Estimates of hourly prices and water availability were} \\
& {\small based on observations made in previous days.} \\ \hline
\end{tabular}
\end{center}
\end{table}

\begin{table}[t!]
\caption{Factor levels for regulatory power.}
\label{tab:waterfactors}
\begin{center}
\begin{tabular}{l|l} \hline
{\bf Level} & {\bf Description} \\ \hline
None & {\small No price changes were allowed.} \\ \hline
Limited & {\small Participants were allowed to change prices no more than} \\
& {\small three times per day (by a single increment).} \\ \hline
Unlimited & {\small Participants were free to change prices as often and as} \\ 
& {\small much as they desired.} \\ \hline
\end{tabular}
\end{center}
\end{table}

\subsubsection{Protocol}
Forty students and research staff (mean age: 26) from Masdar Institute volunteered for the study.  The participants were randomly and uniformly assigned to the same four conditions as in Study~1.  Each participant was taught, via a slide presentation, how to play the game in the assigned condition.  The participant then played the game in a practice scenario in which robot devices made choices randomly.  Finally, the participant played a simulated 30-day game.  Ten trials for both the Adaptive-None and Simple-None conditions were also conducted (no human subjects required).

\subsubsection{Results}

\begin{figure}[t]
\centering
\vspace{-.1in}
\includegraphics[width=2.5in]{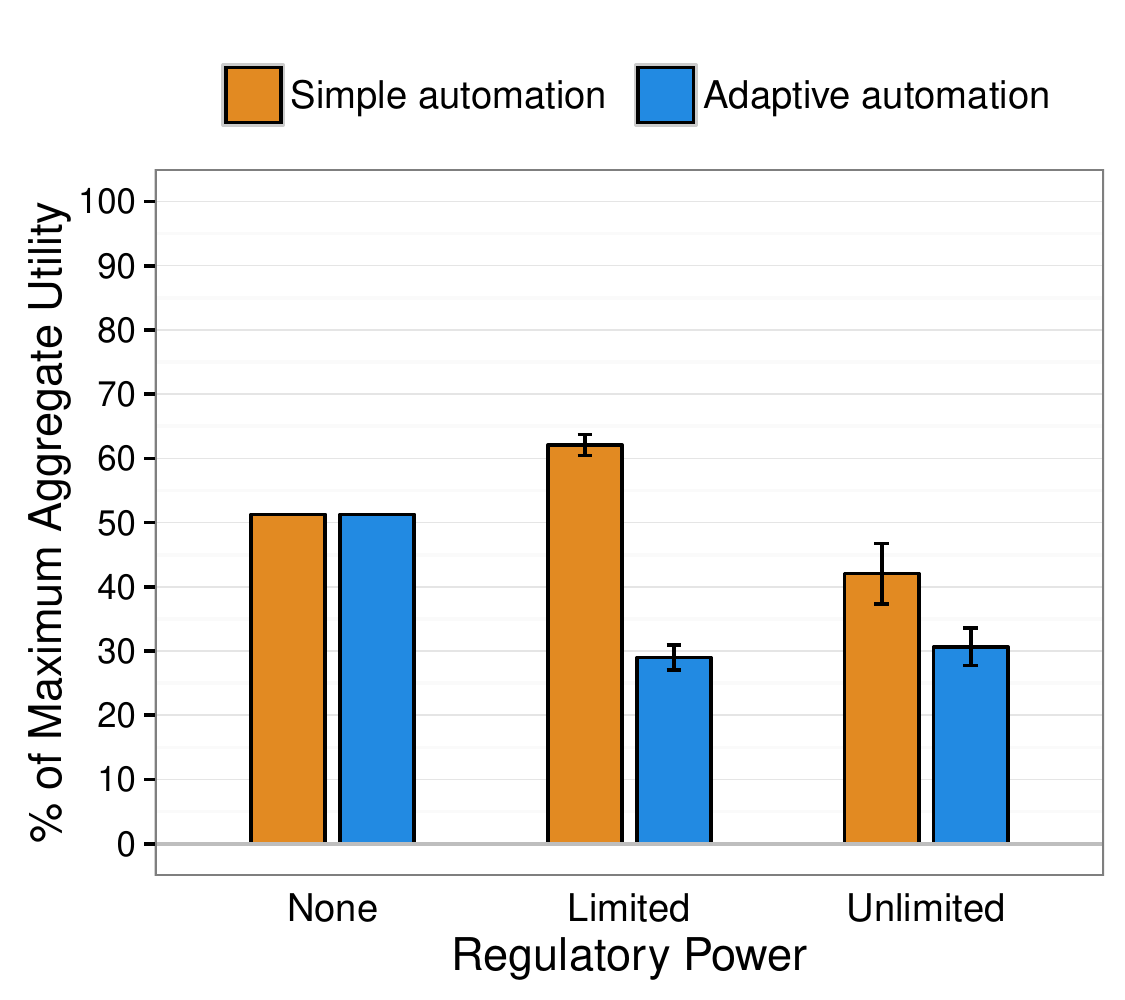} \vspace{-.05in}
\caption{Average aggregate utility during days 26-30 in user study~2 (water management).  Error bars show the standard error of the mean.}
\label{fig:water_utility}
\end{figure}

Participants were asked to set prices so as to maximize the aggregate utility of all tenant's over time.  The average aggregate utility, plotted as a percentage of optimal utility, achieved in each condition is shown in Figure~\ref{fig:water_utility}.  As in Study~1, limited regulatory power produced higher social welfare in the case of simple, non-adaptive, automation.  Unlimited regulatory power likewise produced lower aggregate utility than limited regulatory power when robots used simple automation.  Both limited and unlimited regulatory power led to substantially lower performance when robots used adaptive algorithms.

Statistical analysis confirms these trends.  A two-way analysis of variance, with aggregate utility over the last 5 days as the dependent variable and robot adaptivity and regulatory power as the independent variables, shows a main affect for both regulatory power ($F(1,54)=18.39$, $p < 0.001$), and robot adaptivity ($F(1,54)=53.53$, $p < 0.001$).  There was also a significant interaction affect between robot adaptivity and regulatory power ($F(2,66)=22.91$, $p < 0.001$).  Tukey post hoc analysis shows that, when robots used simple automation, limited regulatory power led to a significant improvement over no regulatory power ($p = 0.037$) and unlimited regulatory power ($p < 0.001$).  Simple-Limited was also statistically better than Adaptive-Limited and Adaptive-Unlimited ($p < 0.001$), and Simple-Unlimited was better than Adaptive-Unlimited ($p = 0.023$).  Finally, any regulation decreased the performance of societies of adaptive robots ($p < 0.001$).


\section{User Study 3 -- Supporting Regulators}
\label{sec:third}

The user studies described in the previous section evaluated two specific HARE.  Interestingly, outcomes from both studies tell a similar story: high regulatory power combined with adaptive robots produced less efficient HARE.  On the surface, these results are counter-intuitive, as both innovations seem to offer {\em more}.  Theoretically, increased regulatory power gives the regulator more leverage over the HARE.  In practice, too much regulatory power appears to have diverted the regulator away from effectively modeling the HARE.  Similarly, adaptive control algorithms allow robots to, theoretically, adapt to each other, thus potentially moving the HARE towards cooperative solutions.  In practice, it appears that the increased complexity of adaptive robots made it more difficult for participants to model (and, thus, influence) these HARE. 

While both high regulatory power and adaptive robot control algorithms failed in the previous two studies, they may add value to the HARE under the right circumstances. One possibility is to assist the regulator in modeling the HARE. Thus, we next consider a third user study in which we gave the regulator automated support in the form of a warning system~\cite{Laughery2006} that forecasted the future state of the HARE, and warned the regulator of potentially undesirable future events.  We again consider the driverless-car scenario used in Study~1.

\subsection{Scenario Overview}

We used a discrete-event simulation (DES) to forecast the future status of each road in the network.  To do this, the system modeled the percentage of cars that chose each road at each node.  These percentages, along with the number of cars currently on each road, were used to simulate the network 20 seconds in advance.  The resulting simulation correctly predicted changes in future system states approximately 80\% of the time.  If the estimated number of cars on a road exceeded the road's capacity at any time during the simulation, then the corresponding road was highlighted in red on the GUI.  Similarly, if the estimated number of cars on a road was between 75-100\%, the road was highlighted in yellow on the GUI.

\subsection{Experimental Setup and Protocol}
The experimental setup and protocol was identical to our first user study, with the exception that participants were warned of pending congestion.  Forty-eight participants (mean age: 28) participated this study.  Twelve subjects were randomly assigned to each condition.

\subsection{Results}
\begin{figure}
\centering
\includegraphics[width=2.3in]{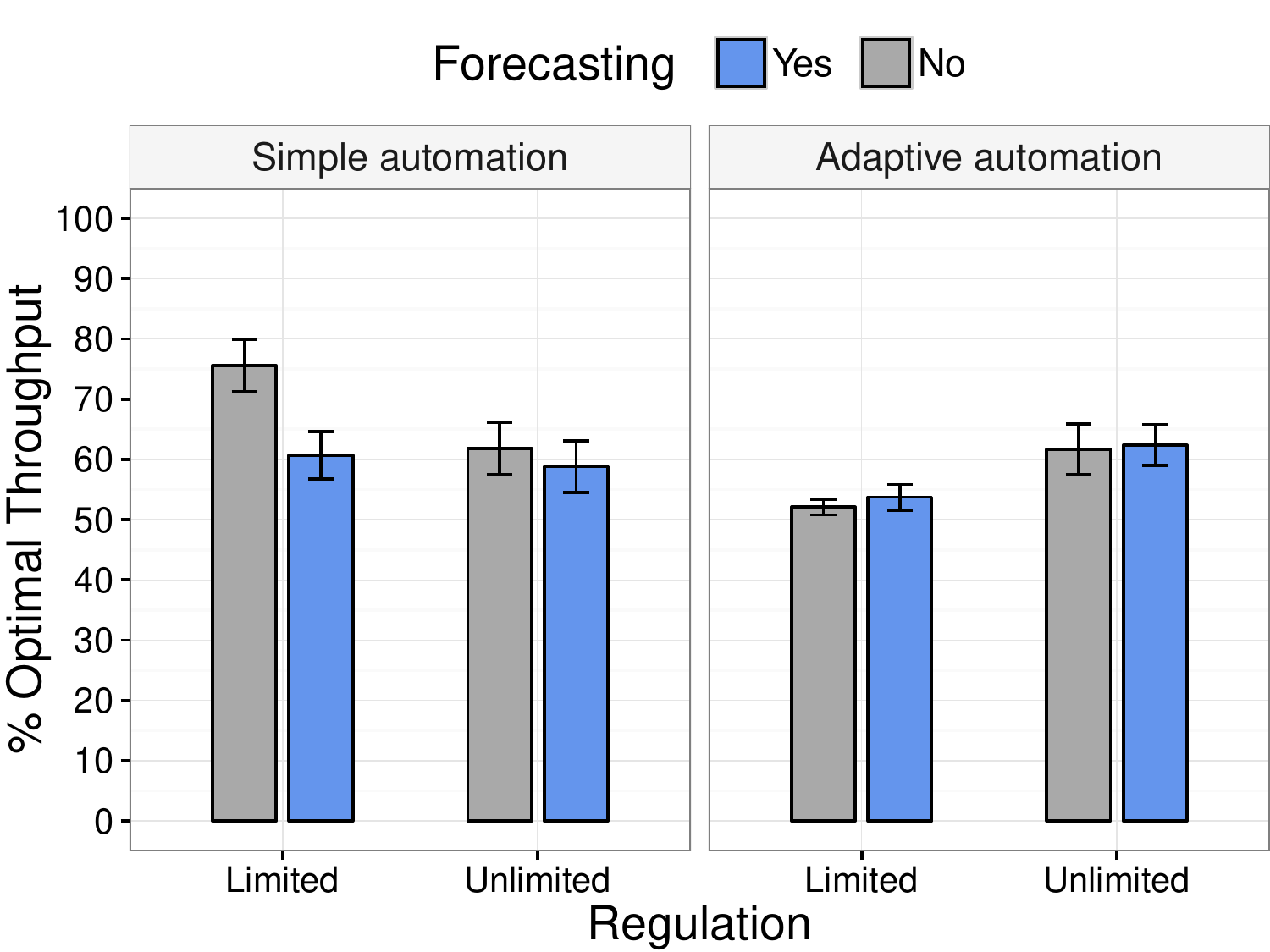}
\caption{System throughput with and without forecasting.}
\label{fig:forecasting}
\end{figure}

Figure~\ref{fig:forecasting} compares the average system throughput obtained when participants were given the warning system verses when they were not.  In most conditions, the decision support system had little impact on the resulting performance of the HARE.  The only exception was in the Simple-Limited condition, where the warning system actually appears to have {\em decreased} throughput.  A two-way independent-samples t-test confirms this observation.  In the Simple-Limited condition, the warning system significantly decreased throughput ($M=-1.31$, $SD=1.81$); $t(21.743)=-2.55$, $p = 0.019$.

The post-experiment questionnaire highlights a potential explanation for the failure of the warning system.  In the Simple-Limited condition, participants given the warning system had a poorer model of the HARE than those that were not given the warning system (Table~\ref{tab:bothright}).  With the warning system, just three of the twelve participants in the Simple-Limited condition correctly identified which node more cars preferred, whereas nine of the twelve participants without the warning system correctly identified the preferred node.

\begin{table}[t!]
\caption{\# of subjects that correctly identified the node preference and vehicle type with (Yes) and without (No) forecasting.}
\label{tab:bothright}
\begin{center} 
\begin{tabular}{l|cc} \hline
 & {\bf Node Preference} & {\bf Vehicle Type} \\ 
{\bf Condition} & No / Yes & No / Yes \\ \hline
Simple-Limited & 9 / 3 & 8 / 6 \\
Simple-Unlimited & 7 / 5 & 1 / 3 \\
Adaptive-Limited & 6 / 7 & 7 / 9 \\
Adaptive-Unlimited & 5 / 5 & 6 / 8 \\ \hline
\end{tabular}
\end{center}
\end{table}

Rather than learning the HARE's tendencies, it appears that some participants instead depended on the forecasting system to identify when congestion was likely to occur.  Since the warning system did not supply instructions for how to alleviate the problem~\cite{Laughery2006}, these participants did not know what to do once a potential problem was identified -- they did not have sufficient knowledge of the HARE's underlying tendencies.  

In short, the decision-support system likewise failed to help the system take advantage of additional regulatory power and adaptive automation, and in fact appears to have made things worse.  The forecasting system identified symptoms of the underlying system, but did not help the regulator model the HARE.  This negative result potentially highlights the role that a decision-support system should play in overcoming the difficulties of adaptive robots and high regulatory power.  We anticipate that decision-support systems for HARE should focus on either helping the regulator to (1)~form an appropriate model of robot behavior or to (2)~balance the time spent modeling and implementing interventions.

\section{System-Specific or General Trends?}
\label{sec:principios}

The results from the user studies reveal somewhat counter-intuitive trends about specific HARE.  In particular, limited regulatory power combined with HARE with simple automation produced the best results.  Would we expect these trends to generalize to other scenarios and systems, including HARE that used different adaptive control algorithms, were regulated by more or less experienced regulators, or that provided the regulator with different user interfaces?  While future work is required to answer these questions in full, we seek to begin to understand the forces that impact the ability of people to regulate HARE.  To do this, we use a simple mathematical model of the regulator to identify the following three general principles that appear to be influential in bringing about the results observed in our user studies.

\vspace{-.16in} ~\\ 
\noindent {\bf Principle 1}: {\em Adaptive robot behaviors typically require the regulator to spend more time modeling the system.}

\vspace{-.16in} ~\\
\noindent {\bf Principle 2}: {\em Adaptive robots typically require the regulator to have higher regulatory power to effectively model the HARE.}

\vspace{-.16in} ~\\
\noindent {\bf Principle 3}: {\em Increased regulatory power tends to decrease the time the regulator spends modeling the HARE.}

\noindent These principles appear to be applicable to all HARE, though the design of the HARE could impact the degree to which they are manifest.  We discuss each in turn.

{\em Principle 1:} To model robot behavior, the regulator must understand how the robots will collectively react to each situation $(s_t, h_t)$, where $s_t$ is the current state of the system at time $t$ and $h_t = (i_0,\cdots,i_{t-1})$ is the intervention history the regulator has implemented up to time $t$.  Here, $i_k$ is the intervention carried out at time $k$.  Let ${\mathcal M}(s,h,i)$ describe how the robots will react when the regulator issues intervention $i$ given system state $s$ and intervention history $h$.

Since the robots' behavior is unknown {\em a priori} to the regulator, the regulator must estimate ${\mathcal M}(s,h,i)$ by observing the robots for each $(s,h,i)$ 3-tuple.  The robot's control algorithms impact the amount of time that must be given to forming the model ${\mathcal M}(s,h,i)$.  In line with neglect benevolence~\cite{NeglectBenevolence}, less time is required to model robots that use stationary decision-making processes than adaptive ones, since adaptive algorithms first adapt to the new intervention, and then react to the reactions of other robots to the intervention, and so on.


Adaptive automation also requires the operator to make more observations than simple automation.  Stationary decision-making processes are typically only contingent on the current system state $s$, whereas adaptive ones are contingent on the tuple $(s, h)$.  Thus, regulators must model a larger state space.


{\em Principle 2:} As discussed for Principle 1, adaptive algorithms require regulators to model the function ${\mathcal M(s,h,i)}$ rather than the simpler function ${\mathcal M(s,i)}$.  Since this model is constructed by observations that require the regulator to implement some intervention $i$, regulators of HARE in which robots use adaptive algorithms must have more regulatory resources to implement the necessary interventions.

{\em Principle 3:} More regulatory power means that regulators (a)~select from a larger set of possible interventions and (b)~have the ability to implement a greater number of interventions.  Having more options can obviously be beneficial, but it comes at the cost of requiring the regulator to spend more time finding the best intervention among all its choices.  Furthermore, implementing a greater number of interventions takes more of the regulator's time (e.g., Figure~\ref{fig:attention_error}a).  Since the regulator must divide its time between modeling the HARE, computing effective interventions, and implementing these interventions, both of these trends mean that more regulator power can reduce the amount of time the regulator spends modeling the system.  This, in turn, can lead to a poorer model of the HARE, as was observed in study 1 (Figure~\ref{fig:attention_error}b).

The forces introduced as Principles 1-3 do not necessarily mean that adaptive control algorithms or more regulator power are always bad.  We anticipate that both developments can still add value when measures are taken to counteract these forces.  Future work should identify how to best do so.

\section{Conclusion}

In this paper, we have presented and discussed the results of three user studies in which people regulated simulated highly automated robot ecologies (HARE).  These studies provide data points that give potential insights into how we can design systems that allow people to regulate HARE so that they meet societal objectives.  Though these data points only provide samples of specific HARE, they highlight easily encountered pitfalls in the design of HARE: seemingly desirable regulatory power, decision support, or adaptive robot autonomy can all lead to HARE with diminished social welfare.  Our results suggest that designers of Human-HARE systems should base design decisions regarding decision support and regulatory power on helping regulators to identify and understand the underlying dynamics of the HARE rather than fixating on controlling current or future system states.  Simultaneously, these data points suggest that designers of HARE should consider limiting the complexity of algorithms used by robots in the HARE, or at least to make the algorithms more immediately transparent to regulators, as simple robot autonomy coupled with limited regulatory power produced the best results.

While illuminating, we must be careful to not overstate the generality of these results, which were obtained for specific (simulated) systems.  Varying any attribute of these systems (e.g., the skill and experience of the regulators; the algorithms, hardware, and information used by the robots; and the communication environment itself) could impact the results.  Our studies are intended to begin to raise awareness of important issues and general principles that should be understood, weighed, and (where necessary) appropriately counteracted as we design real-world HARE.  Future work is needed to better understand, work with, and expound upon these principles.

\appendix

\noindent {\em \bf A. Studies 1 and 3 -- Road Physics:}
Congestion occurred when the number of cars on the road exceeded the road's capacity.  A car's speed on road $ij$ was $V_{ij} \propto [1 / (1 + e^{0.25(N_{ij} - C_{ij})})] + 0.1$, 
where $C_{ij}$ and $N_{ij}$ were the capacity and the current number of cars on road $ij$, respectively.  Thus, as traffic volume reached the road's capacity, traffic flow slowed substantially.  

\vspace{.05in}
\noindent {\em \bf B. Studies 1 and 3 -- Robot Behavior:}
Each simulated car tried to maximize its estimated expected utility, which was based on Eq.~(\ref{eq:utility}).  A new set of destination utilities $v(g)$ for each node $g \in \{A,B,C,D\}$ was generated randomly from a normal distribution each time a car reached its selected destination.  Formally, $v(g) = {\mathcal N}(b(g)+r[0,1], 0.1+0.3(r[0,1])$, where $r[0,1]$ denotes a uniform random selection from the interval $[0,1]$, and $b(g) = 0.6$ for $g \in \{A,B,D\}$ and $b(C) = 0.8$.  This created a preference across the HARE for node C. 

The estimated travel cost $c_\$(i,g)$ (Eq.~\ref{eq:utility}) was the sum of individual link costs along the shortest path to the destination.  Let $c_\$(i,j)$ (defined for adjacent nodes $i$ and $j$) denote the estimated cost for traveling from $i$ to $j$.  Then, for cars using simple automation, $c_\$(i,j) = y \cdot x_{ij}$, where $y=0.079$ was the operating cost per unit time, $x_{ij} = L_{ij}/S_{ij}$, $L_{ij}$ was the length of road $ij$, and $S_{ij}$ was the car's max speed.

Cars employing adaptive automation used reinforcement learning to estimate travel costs.  Initially, $c_\$(i,j)$ was set as in simple automation.  Thereafter, each time a car finished traversing road $ij$, it updated $x_{ij}$ such that $x_{ij} = \alpha x_{ij} + (1-\alpha) z$, where $\alpha \in [0,1]$ was chosen randomly for each car, and $z$ was the observed time to traverse road $ij$.

\vspace{.05in}
\noindent {\em \bf C. Study 2 -- System Properties:}
Each day was divided into six periods, and each tenant had one potential activity per time period.  The water tank refilled at a variable rate throughout the day, such that the water-refill rate was defined by the vector $w = (0, 40, 50, 60, 30, 0)$ (measured in {\em water units}).  Since the consumers wished to consume 300 water units per day in aggregate, demand exceeded supply.  Thus, the regulator needed to learn to set prices, via trial and error, so that water was available when the consumers had high-valued activities, which tended to be at the beginning and ending of the day.

\vspace{.05in}
\noindent {\em \bf D. Study 2 -- Robot Behavior}
Formally, let $L(d,h)$ be the amount of water available to a robotic building on day $d$, period $h$.  Then, for day $D$, hour $H$, the robotic building estimates the water level to be $L'(D,H) = 1/(D-1) \sum_{d=1}^{D-1} L(d,H)$.  Additionally, let $p(d,h)$ be the price of water on day $d$, period $h$.  Then, the tenant estimates the price of water on day $d+1$, period $h$ to be $p^\prime(d+1,h) = p(d,h)$.

After the first day, adaptive automation shifted the tenant's activities in day $d$ so as to maximize expected utility.  If $L^\prime(d,\tau) > s(\tau-t)$, then let $x(d,\tau,t) = \max(0,y(d,\tau,t))$, where $y(d,\tau,t) = v(\tau-t) - s(\tau-t) p^\prime(d,t)$.  Otherwise, $x(d,\tau,t)=0$.  Then, the tenant's schedule is shifted in day $d$ by $t^*(d) = \arg \max_{t \in [0,5]} \sum_{\tau = 1}^6 x(d,\tau,t)$ time periods.



%

\balance{}

\bibliographystyle{SIGCHI-Reference-Format}
\bibliography{bibli}

\end{document}